\documentclass[letterpaper, 10 pt, conference]{ieeeconf}  % Comment this line out if you need a4paper

\IEEEoverridecommandlockouts                              % This command is only needed if 
\usepackage[colorinlistoftodos]{todonotes}

\usepackage{caption}
\usepackage{subcaption}
\usepackage{amsmath}
\usepackage{amsfonts}
\usepackage{flushend}
\usepackage[hidelinks]{hyperref}

\title{\LARGE \bf
DiPPeR: Diffusion-based 2D Path Planner applied on Legged Robots   
}

\author{Jianwei Liu$^{*}$, Maria Stamatopoulou$^{*}$, and Dimitrios Kanoulas\\
\thanks{The authors are with the RPL Lab, Department of Computer Science, University College London, Gower Street, WC1E 6BT, London, UK. {\tt\small \{jianwei.liu.21, maria.stamatopoulou.21, d.kanoulas\}@ucl.ac.uk}}% <-this % stops a space
\thanks{*equal contribution}% <-this % stops a space
\thanks{This work was supported by the UKRI Future Leaders Fellowship [MR/V025333/1] (RoboHike) and the CDT for Foundational Artificial Intelligence [EP/S021566/1].  For the purpose of Open Access, the author has applied a CC BY public copyright license to any Author Accepted Manuscript version arising from this submission.}}

\begin{document}

\maketitle
\thispagestyle{empty}
\pagestyle{empty}

%%%%%%%%%%%%%%%%%%%%%%%%%%%%%%%%%%%%%%%%%%%%%%%%%%%%%%%%%%%%%%%%%%%%%%%%%%%%%%%%
\begin{abstract}
In this work, we present DiPPeR, a novel and fast 2D path planning framework for quadrupedal locomotion, leveraging diffusion-driven techniques. Our contributions include a scalable dataset generator for map images and corresponding trajectories, an image-conditioned diffusion planner for mobile robots, and a training/inference pipeline employing CNNs. We validate our approach in several mazes, as well as in real-world deployment scenarios on Boston Dynamic's Spot and Unitree's Go1 robots. DiPPeR performs on average $23$ times faster for trajectory generation against both search based and data driven path planning algorithms with an average of $87\%$ consistency in producing feasible paths of various length in maps of variable size, and obstacle structure. Website: \href{https://rpl-cs-ucl.github.io/DiPPeR/}{https://rpl-cs-ucl.github.io/DiPPeR/}
\end{abstract}

%%%%%%%%%%%%%%%%%%%%%%%%%%%%%%%%%%%%%%%%%%%%%%%%%%%%%%%%%%%%%%%%%%%%%%%%%%%%%%%%
\section{INTRODUCTION}\label{sec:intro}

% 0) The importance of planning for mobile robot

%Legged robots and quadrupeds in particular represent a formidable example of mobile platforms that display higher navigation capabilities with respect to their wheeled counterpart. Legged platforms offer several advantages over wheeled robots, especially in specific scenarios in which one is required to move on various types of terrain, including uneven or rough surfaces, stairs, and slopes while avoiding obstacles.

Mobile robots, and especially legged ones, have the capacity to evolve into multi-purpose machines, useful in many application scenarios, such as production sites, household services, remote inspection, and disaster search-and-rescue~\cite{Kottege2022}. Path planning is crucial in enabling legged robots to navigate autonomously and effectively complete the attributed tasks in various complex environments. Several studies, e.g.,~\cite{Lau2015, Zhang_2022}, were dedicated towards the development of safe and efficient path planning algorithms, many of which utilize traditional methods such as Rapidly-exploring Random Trees (RRT) and $A^{*}$-based methods~\cite{Kanoulas2019, Suryamurthy2019}. However, such approaches often struggle to effectively handle the complexities and uncertainties associated with real-time sensor inputs~\cite{choudhury2018data}. Efficient and reliable path planning for quadrupeds is an ongoing challenge, with data driven approaches, such as Neural A*~\cite{yonetani2021path} and ViT-A*~\cite{liu2023vit}, showing promising efforts into overcoming the shortcomings of traditional approaches. Learning from demonstration methods using image conditioned Diffusion, have also shown promising results in path planning, applied mainly to manipulators~\cite{chi2023, carvalho2023, janner2022}, however, with minimal literature on their application on quadrupeds. Diffusion policies iteratively infer the action-score gradient, conditioned on visual observations. This allows for expression of multi-modal action distributions and scalability to higher-dimensional output space (allowing the generation of sequence of future actions), and training stability while maintaining distributional expressivity~\cite{chi2023}.

\begin{figure}[t!]
    \centering
       \subfloat[Random Noise initialization]{%
          \includegraphics[height=2.4cm]{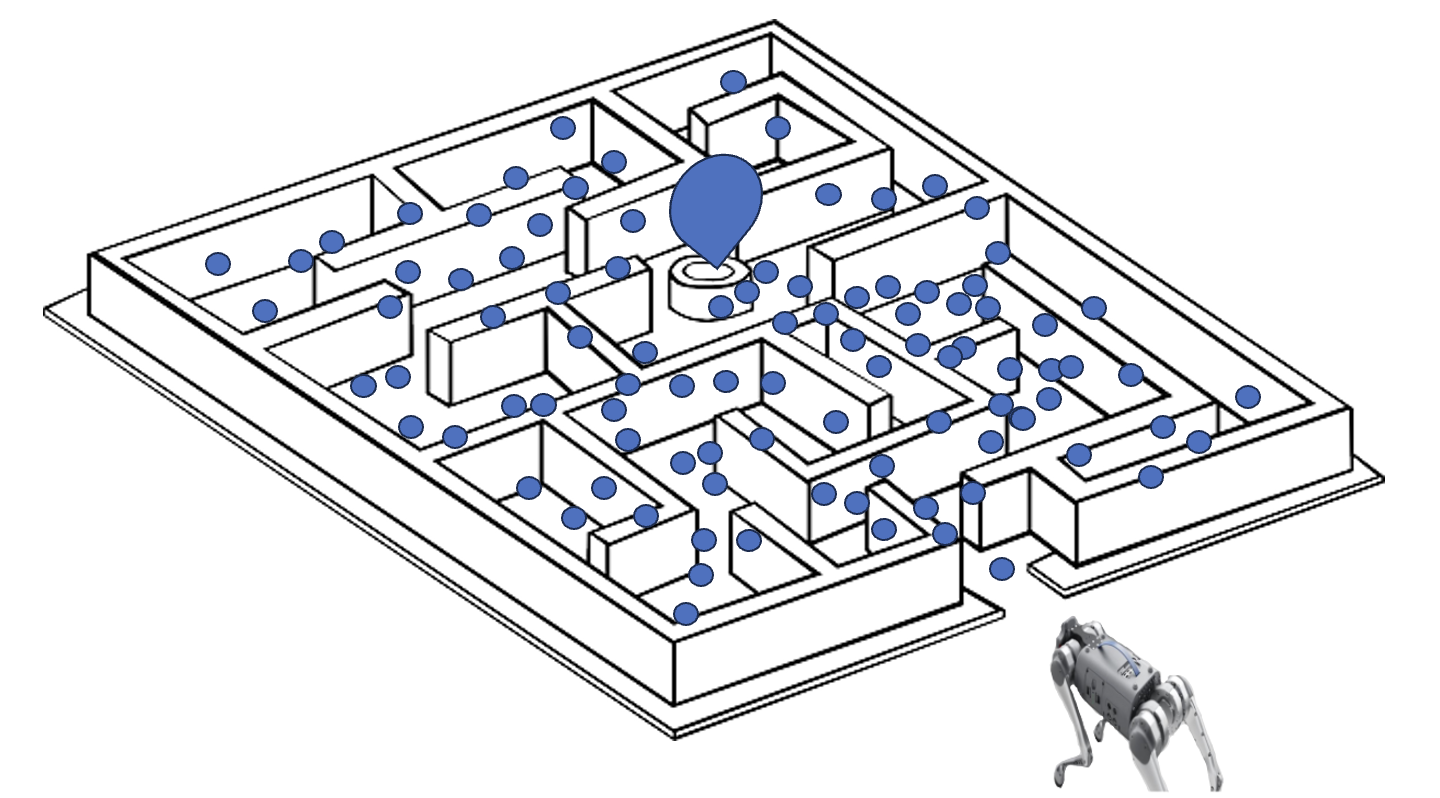}%
          \label{subfig:verynoisypath}%
       } 
       \quad
       \subfloat[Half way through denoising]{%
          \includegraphics[height=2.4cm]{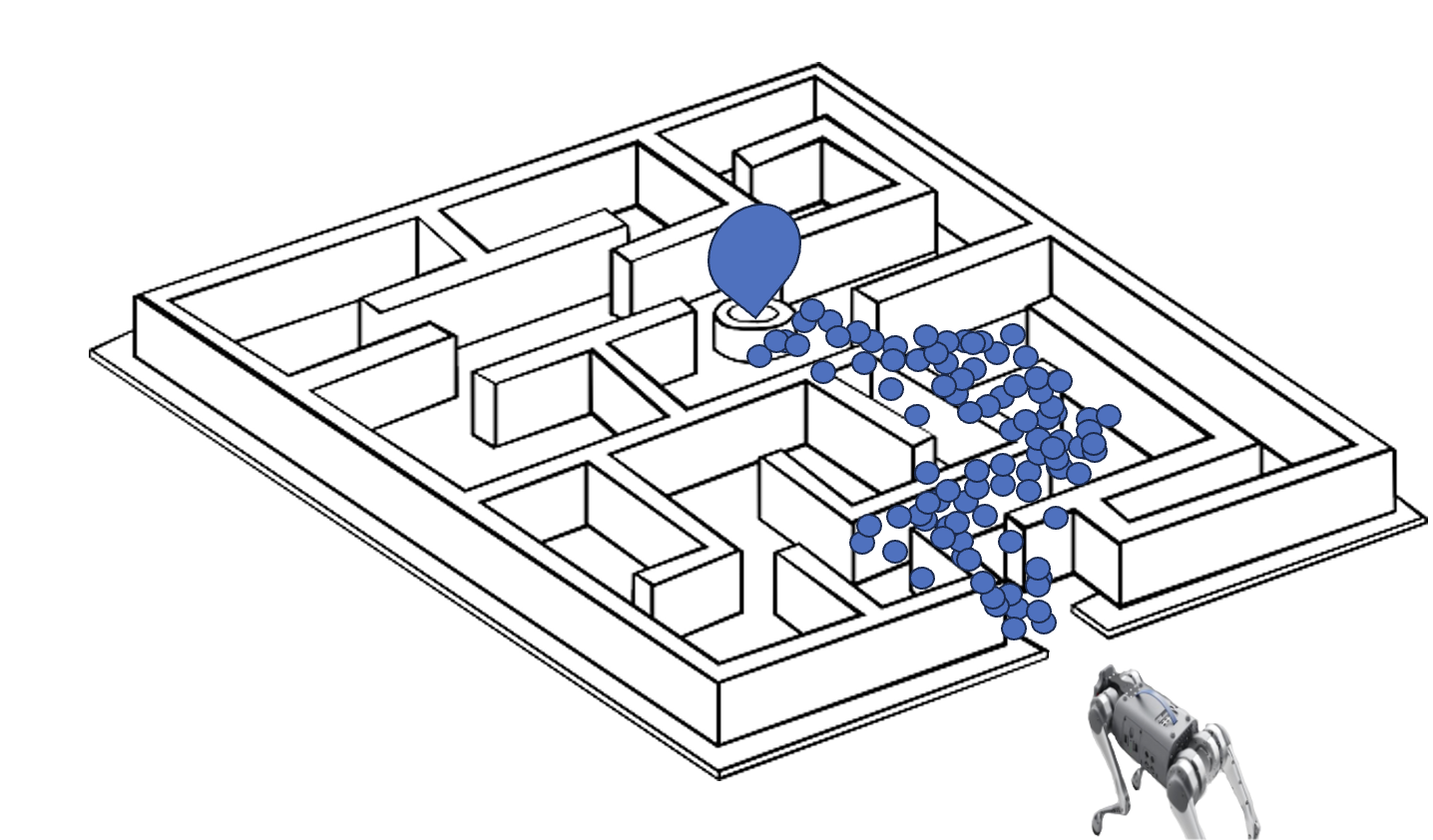}%
          \label{subfig:noisypath}%
       }
       \quad
       \subfloat[Final denoised trajectory plan]{%
          \includegraphics[height=2.4cm]{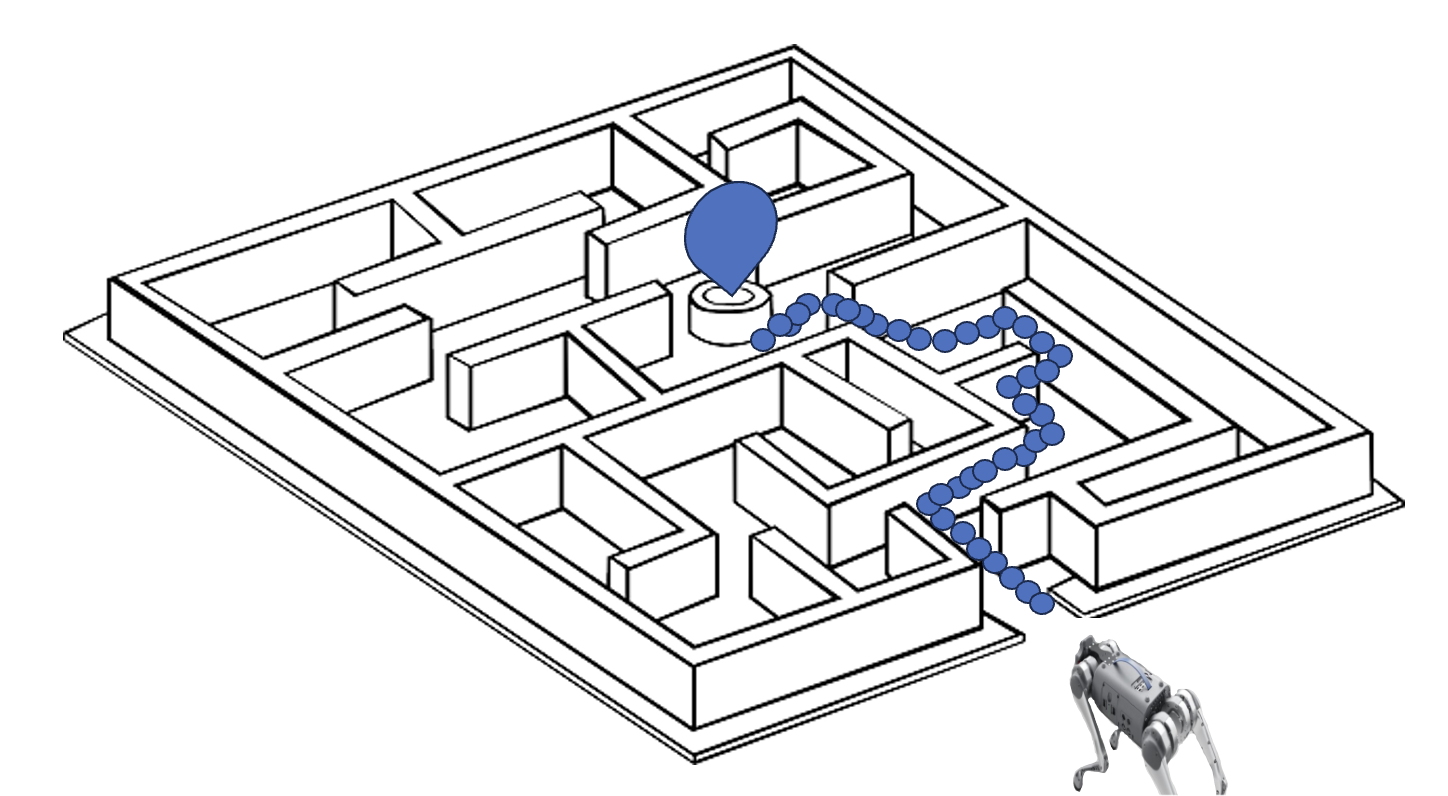}%
          \label{subfig:clearpath}%
        }
    \caption{Illustration of DiPPeR global path generation process. }
    \label{fig:diffglobalpath}
\end{figure}

\begin{figure*}[hbt!]
    \centering\includegraphics[height=5.4cm]{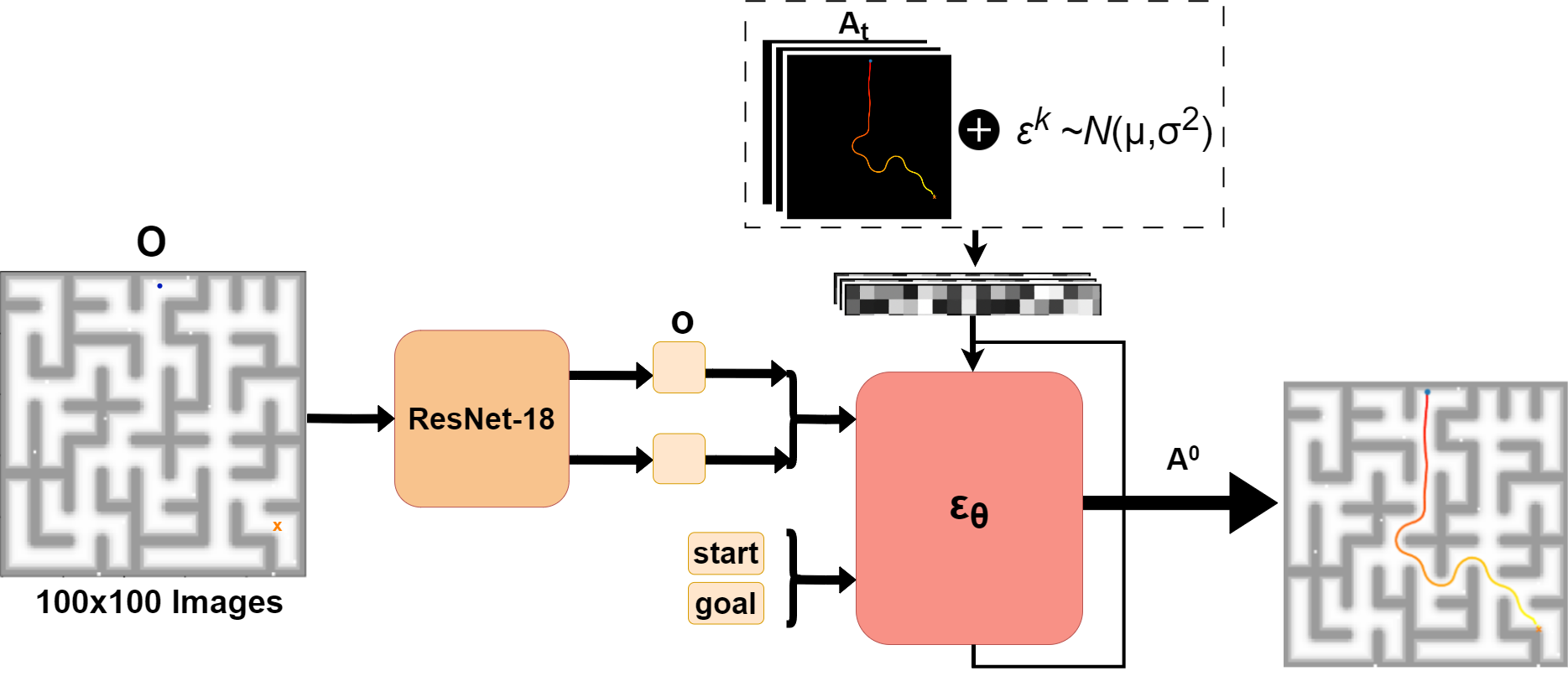}
        \caption{DiPPeR - Image Conditioned Diffusion Training Pipeline: A  Map Image Observations sample $O$ is fed to the ResNet-18 Visual Encoder and converted to latent embeddings $o$. The $x$ and $y$ of the start and goal positions are also added as part of $o$. Noise $\epsilon^{k}$ sampled from the prior Gaussian Distribution is added to the trajectory instance $A_{t}$. The noisy sample is passed as an input to the diffusion network $\epsilon_{\theta}$ and is conditioned by $o$. The network $\epsilon_{\theta}$ takes the form of a CNN and it outputs the denoised action $A^{0}$.}\label{fig:training_pipeline}
\end{figure*}
In this paper, we leverage the progress made in diffusion driven path planning, and develop an 2D path planning framework for quadrupedal locomotion. We develop a training and inference pipeline using a Convolutional Neural Network (CNN) based architecture. The main contributions of our method is the introduction of: 
\begin{enumerate}
    \item a scalable dataset comprising of randomly generated mazes and corresponding trajectories,
    \item an image-conditioned diffusion planner for mobile robots,
    % \item a training/inference pipeline using CNN architecture,
    \item trajectory generation significantly faster than both search based and data driven path planners and
    \item a real-world deployment stack with a platform-invariant framework validation. 
\end{enumerate}

The remaining of the paper is structured as follows. In Sec.~\ref{sec:rw}, we briefly introduce  literature in path planning relevant to our proposed method. In Sec.~\ref{sec:pre}, we provide the necessary background knowledge. In Sec.~\ref{sec:method}, we define our proposed method, with our experimental results presented in Sec.~\ref{sec:exp}. Finally, in Sec.~\ref{sec:conclusion}, we summarize the results and we conclude with some future work.

%(to rewrite)One example of a quadrupedal robot employing RRT for planning can be found in the work of [cite specific paper or research project]. Similarly, another notable instance of a quadrupedal robot utilizing the $A^{*}$ algorithm for planning can be observed in the research conducted by [cite relevant paper or project].

\section{RELATED WORK}\label{sec:rw}
Path planning algorithms, have a long history in robotics and are primarily split between classical and data-driven.

\subsection{Classical Path Planning}
Classical approaches in path planning rely on search-based and sampling-based methods. Search-based path planning provides mathematical guarantees of converging to a solution if it exists.  $A^*$ and its variations, offer simplicity in implementation and effectiveness to find valid paths. For instance, in very recent works~\cite{Huang2022}, the authors introduced an extension of  $A^*$ to drive a mobile platform to sanitize rooms. In~\cite{Kusuma2019}, an $A^*$ algorithm was used to find the collision-free path for a legged robot to achieve autonomous navigation, while in~\cite{sushrutha2021reconfigurable, Ellis2022}, similar path planners were developed for wheeled-legged path planning. For legged robots, the problem is connected to footstep planning too~\cite{kanoulas2017vision, kanoulas2018footstep, kanoulas2019curved}, where a sequence of footsteps are searched for navigation. Traditional methods heavily rely on fixed heuristic functions, such as the Euclidean distance, which lacks adaptability to varying robot configurations and environments and are usually computational heavy. In our work, such heuristics are not required as the path is learned through demonstration of multiple optimal trajectories. 

Sampling-based planners efficiently create paths in high-dimensional spaces, by sampling points in the state space.  Relevant literature in  the field of quadrupedal robots include~\cite{Lau2015}, where an extension of an RRT-based algorithm is used for controlling a quadruped robot during the DARPA Robotics Challenge in 2015. More recently in~\cite{Zhang_2022}, a novel sampling-based planner was introduce to shorten the computational time for finding a new path for quadrupedal robots. While these approaches demonstrate satisfactory performance and probabilistic convergence, their limitations lie in the increasing planning time as the complexity of the environment increases, due to the iterative nature of the algorithms.

\subsection{Data-Driven Path Planning}
State-of-the-art research in the field has shifted towards incorporating machine learning techniques, which directly learn the behavior of path finding. These methods employ approaches such as expert demonstration~\cite{pfeiffer2017perception} or imitation learning~\cite{choudhury2018data} to learn how to plan paths. Recent works directly address the issue of lack of semantically labeled maps in classical search-based methods by using data-driven approaches directly on raw image~\cite{ichter2019robot, choudhury2018data, lee2018gated}. Specifically, Yonetani et al.~\cite{yonetani2021path} introduced Neural $A^*$ (N-$A^*$) -- a differentiable variant of the canonical $A^*$, coupled with a neural network trained end-to-end. The method works by encoding natural image inputs into guidance maps and searching for path-planning solutions on them, resulting in significant performance improvements over previous approaches in terms of efficiency. In an extension of N-$A^*$, Liu et al.~\cite{liu2023vit} introduced ViT-$A^*$, that uses vision transformers for legged robot path planning which further improved the performance. However, as these method still relies on the A$^*$ to generate the final path, these methods would again results in increasing planning time as the complexity of the environment increases.  Our proposed method, utilizes diffusion process to parallelize the generation of the entirety of the trajectory, overcoming this limitation.

% Transformers have emerged as a promising alternative to CNNS, exhibiting significant performance improvements in various computer vision tasks~\cite{codetr2022, wang2023one, Jinjing2023} and robot vision tasks~\cite{ha2022semabs, hao2020prevalent, Hadjivelichkov2022affcorrs}. Transformers have the ability to capture long-range dependencies in images, thanks to their self-attention layers that enable them to attend to any part of the image regardless of the distance from the current location~\cite{dosovitskiy2020image, Raghu2021DoVT}. This is in contrast to CNNs, which are confined to focus on local image patches. Moreover, due to their stacked layers, transformers can learn more complex relationships between different parts of the images while assuming fewer inductive biases~\cite{Raghu2021DoVT}.

\begin{figure*}[hbt!]
\centering
   \subfloat[$k=0$]{%
    \includegraphics[height=2.cm]{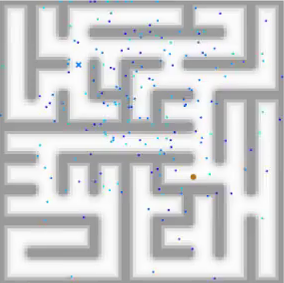}
      \label{subfig:it0}% 0.243
      } 
   \subfloat[$k=300$]{%
     \includegraphics[height=2.cm]{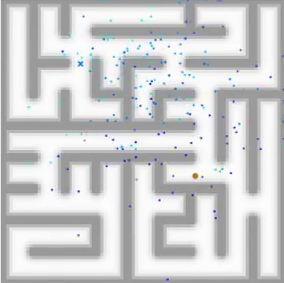}
      \label{subfig:it250}%
      } 
   \subfloat[$k=700$]{%
     \includegraphics[height=2cm]{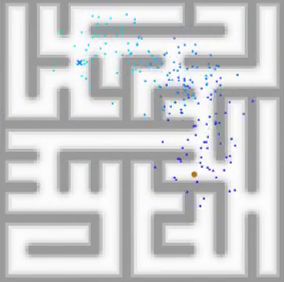}
      \label{subfig:it500}%
      } 
   \subfloat[$k=1000$]{%
      \includegraphics[height=2.cm]{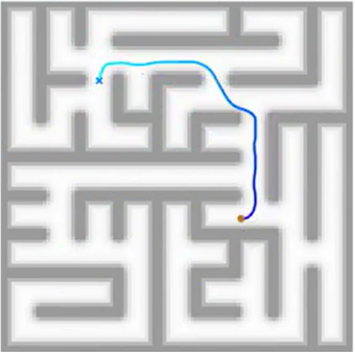}
      \label{subfig:it1000}%
      }

    \caption{Denoising diffusion steps ($k=1000$, path$_{l}=200$) to generate a path from noisy samples.}
    \label{fig:denoising_path_example}
\end{figure*}

\subsection{Diffusion for Path Planning}
Diffusion methods have gain popularity in the domain of path planing with many works presenting promising results. Hong et al.~\cite{Hong2018}  developed diffusion maps applied to find a local path for reaching a goal and avoiding collisions with dynamic obstacles simultaneously, by computing transition probabilities between grid points. Janner et al.~\cite{janner2022} and Chi et al.~\cite{chi2023} developed impressive path planners, applied to robotic manipulators, by providing demonstration data and learning the trajectories through diffusion. We aim to leverage the promising results and develop a diffusion planing pipeline applied to quadrupedal locomotion.  
\section{Preliminaries}\label{sec:pre}
Path planning is essential for robot autonomous navigation and involves calculating a trajectory for a robot to follow in a map, between a start and end point. The solution of path planing refers to the generation of an optimal path, that full-fills the properties of finding the collision free, shortest, and smooth route between start and goal positions~\cite{Tzafestas2014}.  As elaborated in Sec.~\ref{sec:rw}, there are plethora of approaches to solve path planing. For relevance to our method we expand on Probabilistic Diffusion-based path planing, leveraging the learning capabilities of CNNs.

\subsection{Diffusion}
Image-guided diffusion models~\cite{hazart2023} have emerged as a powerful generative model with impressive performance when dealing with image datasets, among others. They provide the ability to transform a latent encoded representation into a more meaningful description of the image data. A popular variation is the Denoising Diffusion Probabilistic Model (DDPM), a generative model defined through parameterized Markov chains trained using variations inference. A forward chain converts input data into noise and a reverse chain converts the noisy data back to its original form. In particular, the noisy data $x^{1:N}$ is generated by iteratively adding Gaussian noise to the data $x^0$ according to a variance schedule $\beta^{1:N}$~\cite{ho2020denoising}:
\begin{equation}
    q(x^i\mid x^{i-1}) = \mathcal{N}(x^i;\sqrt{1-\beta^i}x^{i-1},\beta^i\mathbf{I})
\end{equation}
Then, the denoising occurs by learning transition kernels parameterized using deep neural networks, for reversing the noisy data $x^{N}$ back to the input $x^0$~\cite{ho2020denoising, yang2022}. The learned denoising kernel $p_{\theta}(x^{i-1}\mid x^{i})$ is parameterized by a prior Gaussian distribution starting at $p(x^{N}) = \mathcal{N}(x^{N};\mathbf{0},\mathbf{I})$ and is defined by:
\begin{equation}\label{diffusion_kernel}
    p_{\theta}(x^{i-1}\mid x^{i}) = \mathcal{N}\left(x^{i-1} ; \mathbf{\mu}_{\theta}(x^{i},i),\mathbf{\Sigma}_{\theta}(x^i,i)\right)
\end{equation}
where, $\theta$ represents the model parameters, $\mathbf{\mu}_{\theta}(x^{i},i)$ the mean and $\mathbf{\Sigma}_{\theta}(x^i,i)$ the variance, parameterized by the deep neural network. A popular deep neural network choice for image conditioned diffusion are CNNs due to their benefits in dealing with image datasets.

\section{METHOD}\label{sec:method}
Our method is inspired by the baseline papers~\cite{chi2023, janner2022}. We adapt the image conditioned diffusion pipeline~\cite{chi2023} to solve the problem of mobile robot path planing, while also conditioning for the starting and goal position of the trajectory via inpainting based on~\cite{janner2022}. 
Initially the training dataset is generated, comprising of the 100x100 sized random and solvable maps and a number of trajectories for each map (Sec.~\ref{data_gen}). This is tenfold larger and more complicated dataset than the one provided in the baseline paper~\cite{janner2022}. The dataset is fed into the training pipeline (Sec.~\ref{subsec:training}), where the optimal trajectories are learned through demonstration by preserving local consistencies. The inference pipeline (Sec.~\ref{subsec:inference}) allows the generation of the optimal trajectory given start and goal positions and the relevant map. 

\begin{figure}[b!]
    \centering
       \subfloat{
          \includegraphics[height=1.7cm]{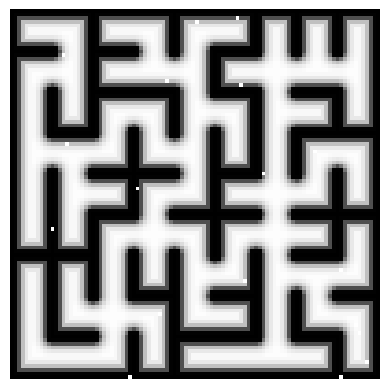}
        }
        \subfloat{
          \includegraphics[height=1.7cm]{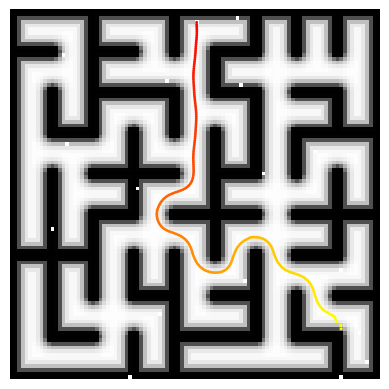}
        }

       \subfloat{
            \includegraphics[height=1.7cm]{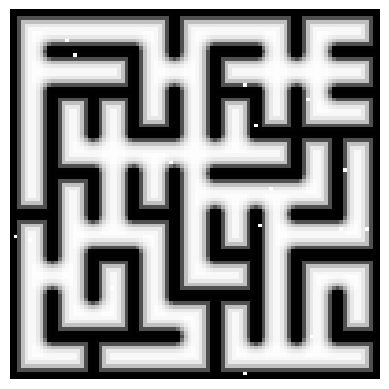}
        } 
        \subfloat{
          \includegraphics[height=1.7cm]{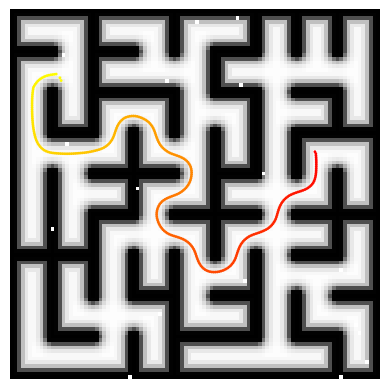}
        }

        \setcounter{subfigure}{0}
        \hspace{0.1cm}\subfloat[Maps]{
              \includegraphics[height=1.7cm]{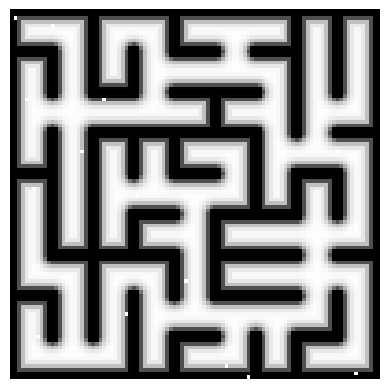}
            \label{subfig:map}
        }\hspace{-0.25cm}
        \subfloat[Trajectories]{
          \includegraphics[height=1.7cm]{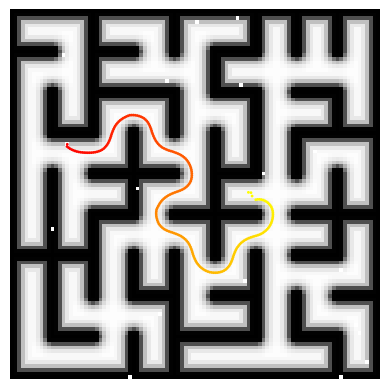}
          \label{subfig:trajectory}
        }

    \caption{Generated samples from the dataset: \ref{subfig:map}) examples of $100\times100$ random solvable maps and \ref{subfig:trajectory}) examples of trajectories, generated through $A^{*}$.}
    \label{fig:obs}
\end{figure}

\subsection{Data generation}\label{data_gen}
 Creating the training dataset includes generating random map images and feasible 2D trajectories. Examples of the randomly generated maps and trajectories are depicted in Fig.~\ref{fig:obs}.

\begin{figure*}[hbt!]
\centering
   \subfloat[]{%
    \includegraphics[height=2.1cm]{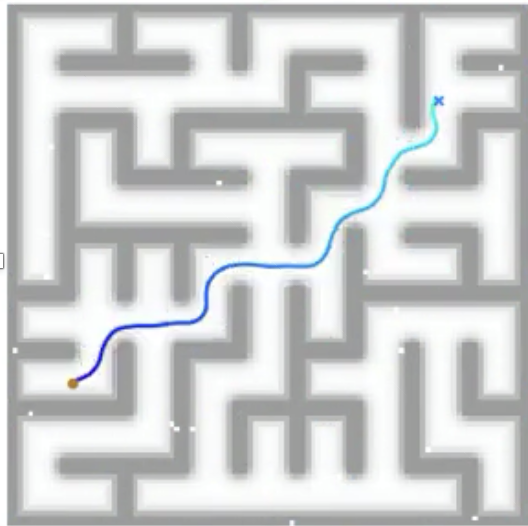}
      \label{subfig:traj1}% 0.243
      } 
   \subfloat[]{%
     \includegraphics[height=2.1cm]{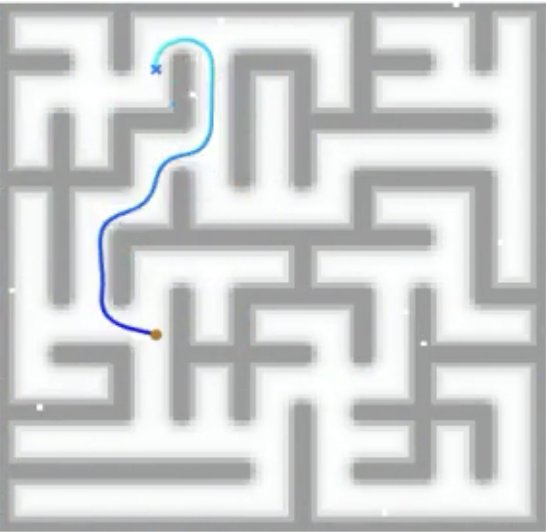}
      \label{subfig:traj2}%
      } 
   \subfloat[]{%
     \includegraphics[width=0.16\textwidth, height=2.1cm]{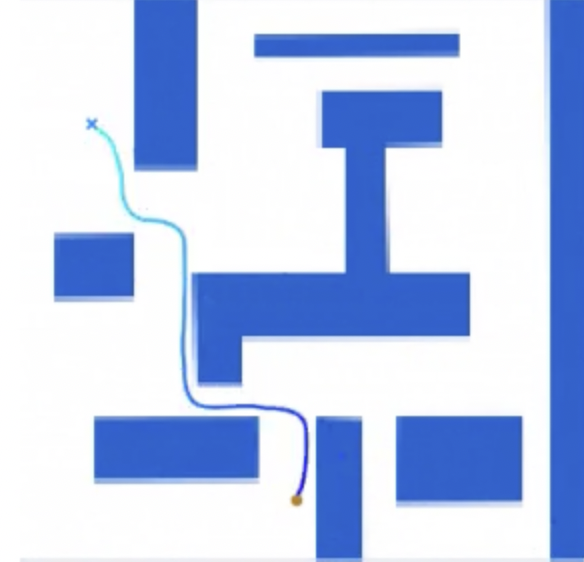}
      \label{subfig:traj3}%
      } 
   \subfloat[]{%
      \includegraphics[height=2.1cm]{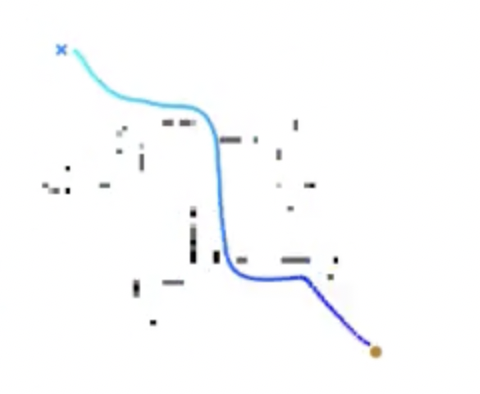}
      \label{subfig:traj4}%
      }
    \subfloat[]{%
      \includegraphics[height=2.1cm]{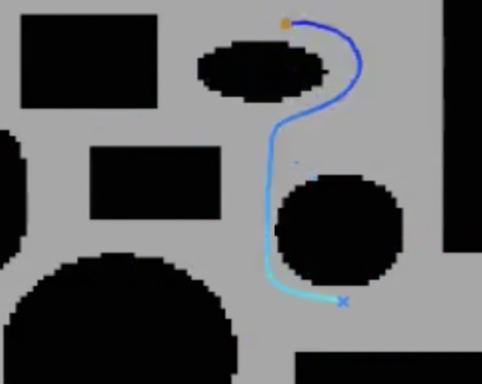}
      \label{subfig:traj6}%
      }
    \caption{Validating DiPPeR's performance and generalization. A random point is selected for the start and goal position on the provided map. Maps a) and b) are part of the validation dataset to validate the performance of the network in connecting the starting and end points while also avoiding the obstacles. Maps c),d) and e) are used to test the ability of the network to generalize to different out-of-distribution environments of varying scale, color and obstacle structure.}
    \label{fig:resultfigures}
\end{figure*}

\subsubsection{Map generation}
Map generation is done through Kruskal’s Minimum Spanning Tree (MST) Algorithm~\cite{kruskal1956shortest}, with the edges representing potential wall locations and the nodes representing cells. The algorithm works by initially considering all edges of a randomly weighted graph and sorting them by their weights. Then, it iteratively adds edges to the MST, starting with the smallest weight, while ensuring that the graph remains acyclic. This process continues until all vertices are in the MST or the desired number of edges is reached. Kruskal's algorithm employs disjoint-set data structures to efficiently detect and avoid creating cycles during edge selection, resulting in a tree that spans all vertices with the minimum possible total edge weight.
\subsubsection{Path generation}
To provide trajectories in the training framework, the generate paths need to be \emph{feasible} - avoid all obstacles. 
To generate feasible paths we use the $A^{*}$ path finding algorithm that uses a combination of heuristic estimates and cost information to efficiently find the shortest path between the start and end nodes. We randomize the position of the start and goal node to generate a variety of trajectories per map. 

\subsection{Training Framework}\label{subsec:training}
% \begin{figure*}[hbt!]
%     \centering\includegraphics[width=0.75\textwidth,height=5.2cm]{Figures/4_method/training_pipeline.png}
%     \caption{DiPPeR - Image Conditioned Diffusion Training Pipeline: Sample from th Map Image Observations $0$ is fed to the ResNet-18 Visual Encoder and converted to latent embeddings $o$. Noise $e_{k}$ sampled from the prior Gaussian Distribution is added to the trajectory instance $A_{t}$. The noisy sample is passed as an input to the diffusion network $\epsilon_{\theta}$ and conditioned is conditioned by $O$. The network $\epsilon_{\theta}$ takes either the form of a CNN or a DDPM-Transformer and it outputs the denoised action $A_{0}$.}
%     \label{fig:training_pipeline}
% \end{figure*}
We formulate DiPPeR as a vision-guided mobile robot path planner generated through DDPMs, as presented in Sec.~\ref{sec:pre}. Our observation space $O$ comprises of the $100\!\times\!100$ pixel images, representing the randomly generated maps. Our action space $A_{t}$ comprises of 2D trajectories for each map. The subscript $t$ refers to a sequence of timesteps, with $t=T$ representing the training horizon.  We empirically conclude that a dataset of $10,000$ maps with $100$ trajectories each, is sufficiently large to achieve generalization and adaptability to unseen maps. Fig.~\ref{fig:training_pipeline} provides a graphical representation of the proposed training framework.

\subsubsection*{\textbf{DDPM Training}}
DDPM in our method, is used to approximate the conditional distribution $p(A_{t}\mid O)$ of the action vector $A_{t}$, given the map image observation $O$. This formulation speeds up the diffusion process and improves the generated actions by predicting an action conditioned to an observation, which translates to predicting trajectories given the specific map image. 

DDPM takes as input $A_{t}^{k}$ with added noise $\epsilon^{k}$, sampled from the prior Gaussian distribution and performs $k$ denoising iterations ($A_{t}^{k-1},A_{t}^{k-2},...,A_{t}^{0}$) through gradient descent, following  Eq.~\ref{eqt:denoising}. The output is the noise free representation of the input vector $A_{t}^{0}$:
\begin{equation}\label{eqt:denoising}
    A_{t}^{k-1} = \alpha(A_{t}^{k}  - \gamma\epsilon_{\theta}(O,A_{t}^{k} ,k) + \mathcal{N}(0,\sigma^{2}I)),
\end{equation}
where $\epsilon_{\theta}$ represents the noise prediction network. The variables $\alpha,\gamma,\sigma$ and $\epsilon^{k}$, when expressed as functions of $k$ compose the noise schedule that drives the learning process, in this case, we've used the Square Cosine Schedule~\cite{nichol2021improved} in line with~\cite{chi2023}. The hyperparameters $\alpha,\gamma,\sigma$ determine the scheduling learning rate which controls the extent to which the diffusion policy captures high and low-frequency characteristics of action signals.

Training $\epsilon_{\theta}$, involves predicting the noise added to a random sample $A_{t}^{0}$, through  Eq.~\ref{eqt:denoising}. For each $A_{t}^{0}$ a denoising iteration $k$ is selected with an added corresponding noise value $\epsilon^{k}$ and variance. The mean squared error between the $\epsilon^{k}$ and the predicted noise value from $\epsilon_{\theta}$ is then calculated based on Eq.~\ref{eqt:mse_noise}, with the aim to be minimized along the gradient descent. 
\begin{equation}\label{eqt:mse_noise}
    \mathcal{L} = MSE(\epsilon^{k}, \epsilon_{\theta}(O,A_{t}^{0} +\epsilon^{k},k))
\end{equation}

% Unlike~\cite{chi2023}, we do not perform receding predictive horizon and do not implement a controller to translate the policy's output action to robot action.
By using inpainting-based goal state conditioning~\cite{janner2022} and image conditioned diffusion~\cite{chi2023}, DiPPeR actions can be directly implemented to the real robot to find feasible trajectories connecting the start and goal positions given the map of the environment. 
We observe that varying the horizon length during training has a significant impact in the performance of the model. The generated trajectories in our dataset have variable length from $10$ to $400$, according to the A* generated path. The horizon length should be long enough to capture the whole range of the dataset, hence we set it equal to $180$ which is the estimated average trajectory length. 

 An important design choice is selecting the architecture of $\epsilon_{\theta}$. We chose a CNN due to it being relative easy to tune compared to Transformers.
 
\subsubsection*{\textbf{DiPPeR}}\label{sec:train_cnn}
We develop two variations of DiPPeR$_{cnn}$. The first version has observation space $O$ as defined in Sec~\ref{subsec:training}. The second version adds two extra terms to $O$, the trajectory start and end points, each expressed as a 2D vector corresponding to the $x$ and $y$ pixel coordinates. Whilst these extra start and goal conditions are not strictly necessary, as the inpainting-based start and goal conditioning are also used during inference, we have noticed that these extra conditions helped with convergance speed during training. A 1D temporal CNN is used with conditioning the actions generation on the observations by $p(A_{t}\mid O)$. The conditioning occurs through Feature-wise Linear Modulation (FiLM)~\cite{perez2018film} as proposed in~\cite{chi2023}.

\begin{figure*}[t!]
\centering
   \subfloat[office01\\$280 \times 280$]{%
        \includegraphics[width=0.15\textwidth]{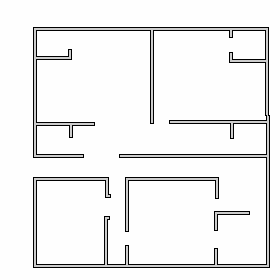}
      \label{subfig:office01}} 
       \hfill
   \subfloat[room02\\$360 \times 360$]{%
     \includegraphics[width=0.15\textwidth]{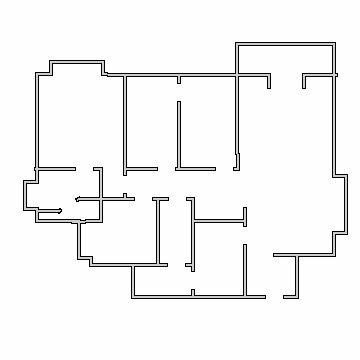}
      \label{subfig:map_room02}} 
       \hfill
   \subfloat[office02\\$600 \times 600$]{%
     \includegraphics[width=0.15\textwidth]{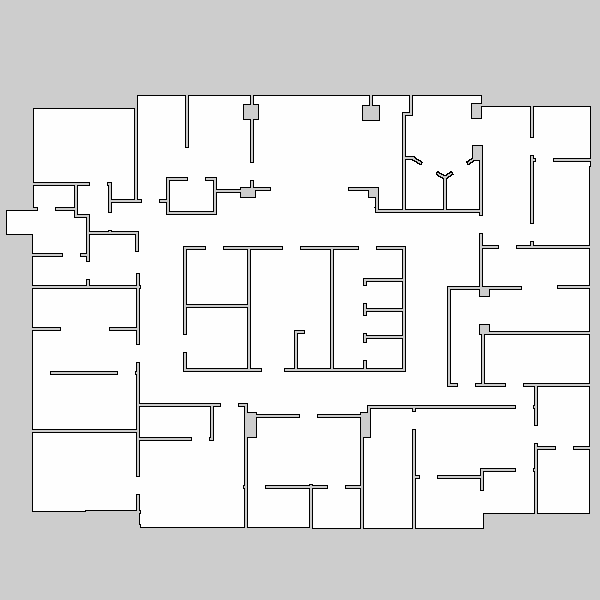}
      \label{subfig:map_office02}} 
       \hfill
   \subfloat[mall\\$760 \times 760$]{%
     \includegraphics[width=0.15\textwidth]{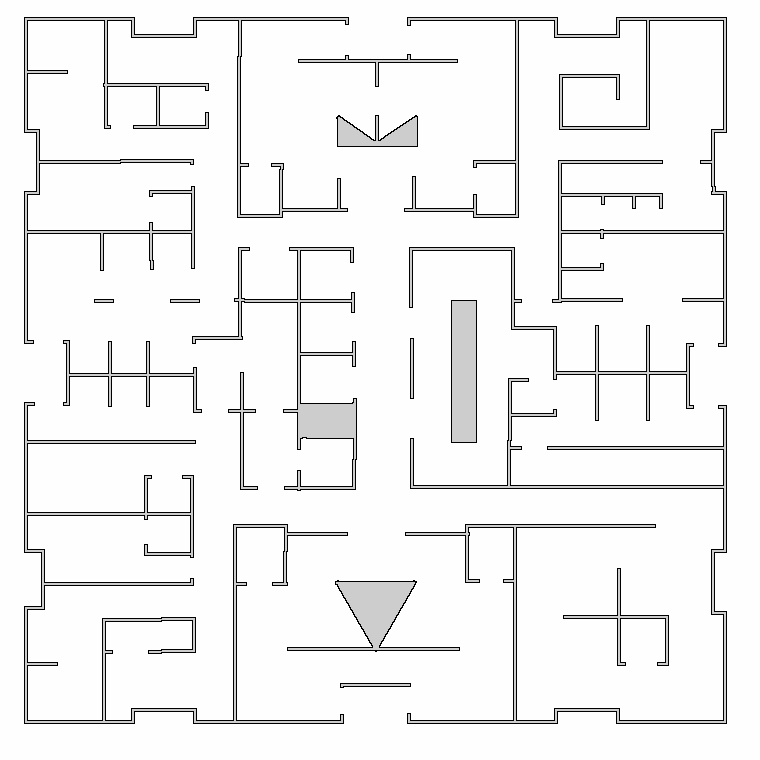}
      \label{subfig:map_maze} } 
    \caption{MRPB benchmark dataset maps used for comparing different planning methods with their respective sizes.} 
    % The blue depicted line corresponds to the DiPPeR generated trajectory.}
    \label{fig:maps_bench}
\end{figure*}

\subsubsection*{\textbf{Visual Encoder}}
A ResNet-18 visual encoder with spacial softmax pooling is trained end-to-end to convert the observation image $O$ to a latent embedding $o$ while preserving spatial information. 
The ResNet is trained alongside $\epsilon_{\theta}$.

% A noisy $A_{t}$ without the first (start) and last (goal) elements, with length equal to $path_{l}$ and added noise $e_{k}$ is fed in the DiPPeR.

\section{RESULTS}\label{sec:exp}
% \begin{figure*}[hbt!]
% \centering
%    \subfloat[$k=0$]{%
%     \includegraphics[height=3.cm]{Figures/5_experiments/Diffusion_denoising_ex_0.png}
%       \label{subfig:it0}% 0.243
%       } 
%    \subfloat[$k=300$]{%
%      \includegraphics[height=3.cm]{Figures/5_experiments/Diffusion_denoising_ex_1.png}
%       \label{subfig:it250}%
%       } 
%    \subfloat[$k=700$]{%
%      \includegraphics[height=3.cm]{Figures/5_experiments/Diffusion_denoising_ex_2.png}
%       \label{subfig:it500}%
%       } 
%    \subfloat[$k=1000$]{%
%       \includegraphics[height=3.cm]{Figures/5_experiments/Diffusion_denoising_ex_3.png}
%       \label{subfig:it1000}%
%       }

%     \caption{Denoising diffusion steps ($k=1000$, path$_{l}=200$) to generate a path from noisy samples.}
%     \label{fig:denoising_path_example}
% \end{figure*}

\subsection{Inference Pipeline}\label{subsec:inference}
After training, the inference pipeline is used to validate DiPPeRs' performance. A start and goal position are randomly sampled from a uniform distribution and are then passed into $\epsilon_{\theta}$, alongside $O$. A $path_{l}$ variable defines the number of noisy samples used during the denoising diffusion process. The value of path$_{l}$ is defined as function of the approximate length of the estimated trajectory, i.e. for start and goal positions being further apart, the $path_l$ will be larger and vice-versa. A vector of noise sampled from the prior Gaussian distribution with length equal to path$_{l}$ and $2$ dimensions (x and y pixel coordinates) is constructed with the inpainting conditioning (i.e. the first and last column of the noise vector being set to the start and goal positions, respectively), is fed in the DiPPeR. The reverse chain of the DDPM model is used to iteratively denoise the input vector. The output is the final trajectory $A_{0}$, connecting the start and goal points, while aiming to follow a feasible path. The progress of denoising during inference is depicted in Fig.~\ref{fig:denoising_path_example}.

\subsection{Simulation Results}
The evaluation of DiPPeR's performance is completed in two stages. \emph{Performance} evaluation is performed by sampling map images from the validation dataset and \emph{Out-of-distribution} evaluation which is performed by selecting unseen map images of varying scale, color and obstacle structure to test DiPPeR's generalization capabilities. 
In both cases a random start and goal position is selected along the map and inference is performed following Sec.~\ref{subsec:inference}. For all experiments the number of diffusion iterations is empirically chosen as $k=1000$, to ensure both full convergence and minimum inference time. In most cases convergence is achieved with a smaller $k$ however we decide to keep it constant to preserve uniformity across experiments. The path length path$_{l}$ parameter has a significant impact on the diffusion performance and needs to be varied according to the desired trajectory length. Given the start and goal position and the structure of the map, the number of feasible pixels can be measured to create an approximate estimation of path$_{l}$. Significantly larger path$_{l}$ will results in  trajectories that loops locally and significantly smaller path$_{l}$ result in trajectories going through obstacles to connect the start and end goal. 
\begin{figure*}[hbt!]
\centering
   \subfloat{%
    \includegraphics[height = 2.1cm]{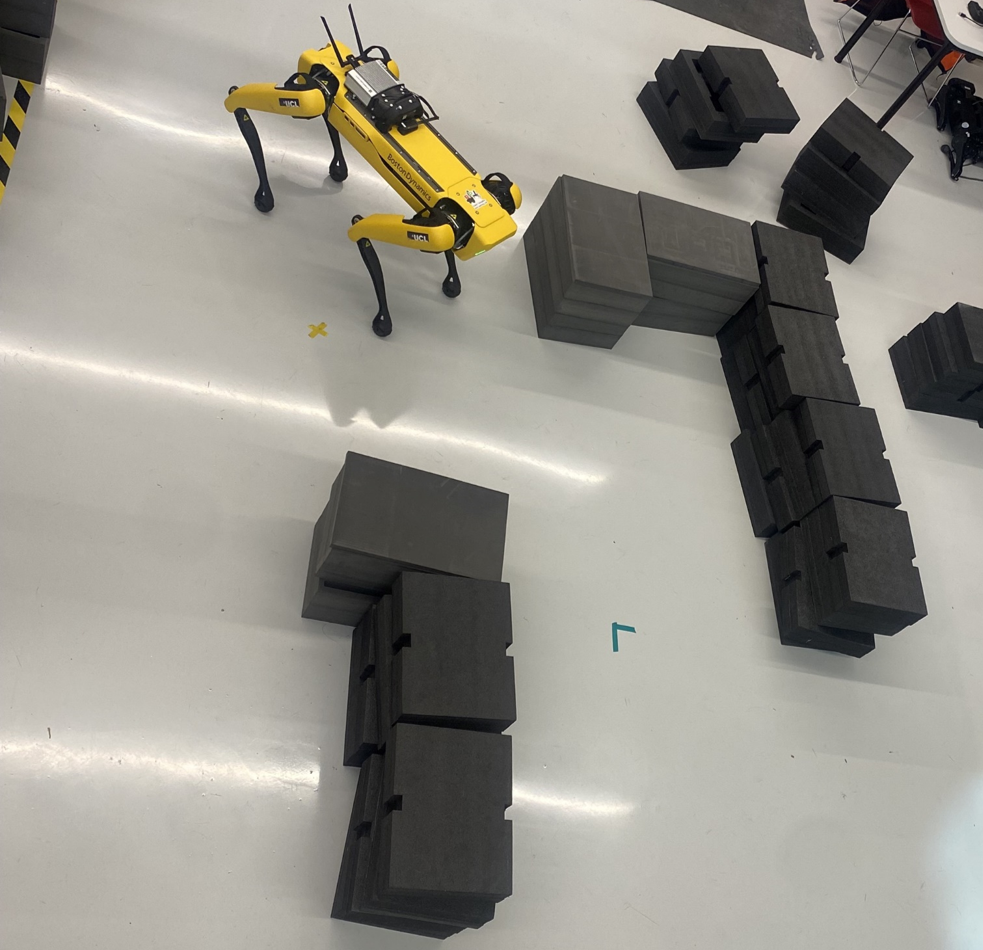}
      \label{subfig:spot_nav_around_obstacle_1}% 0.243
      } 
   \subfloat{%
     \includegraphics[height = 2.1cm]{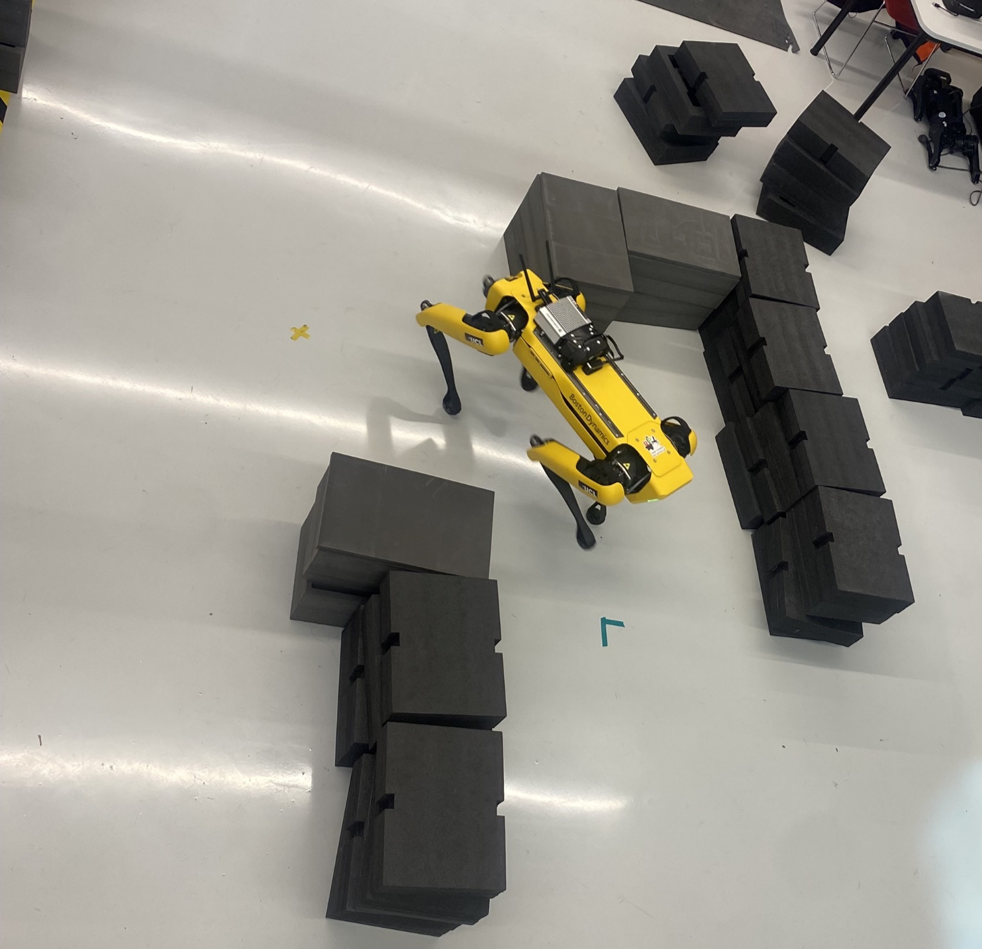}
      \label{subfig:spot_nav_around_obstacle_2}%
      } 
   \subfloat{%
     \includegraphics[height = 2.1cm]{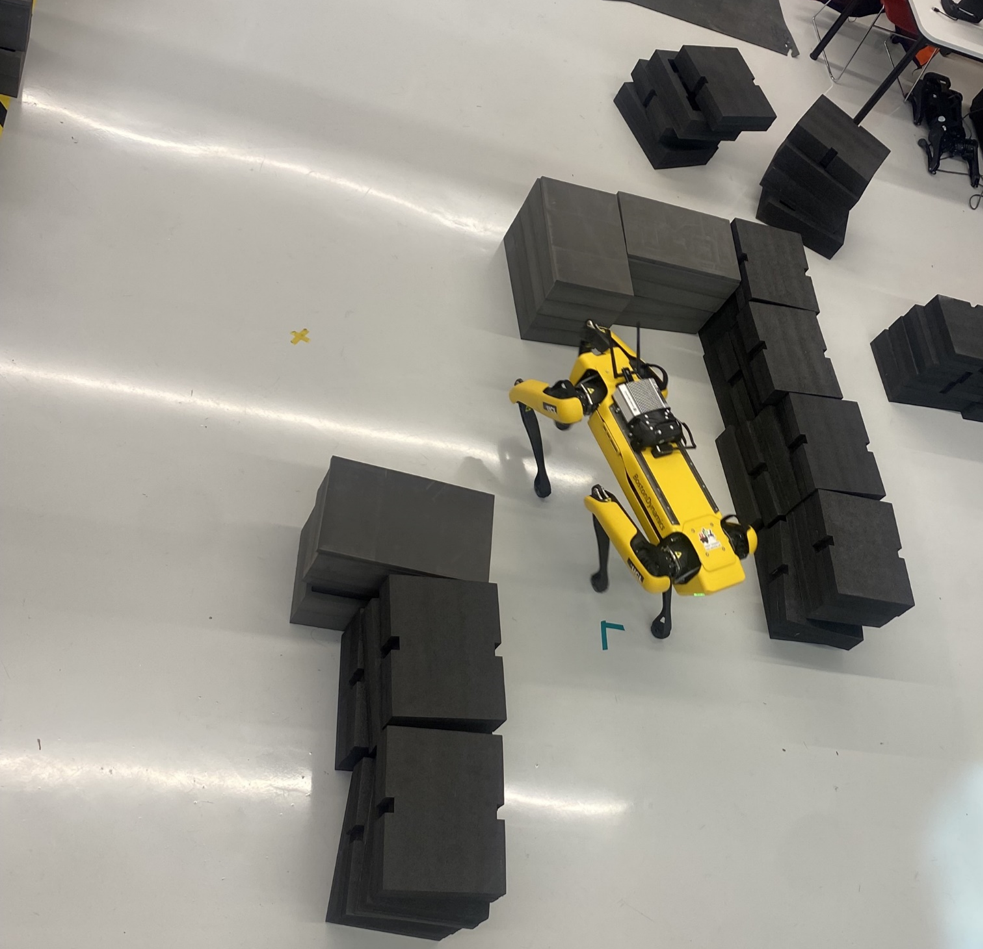}
      \label{subfig:spot_nav_around_obstacle_3}%
      } 
   \subfloat{%
      \includegraphics[height =2.1cm]{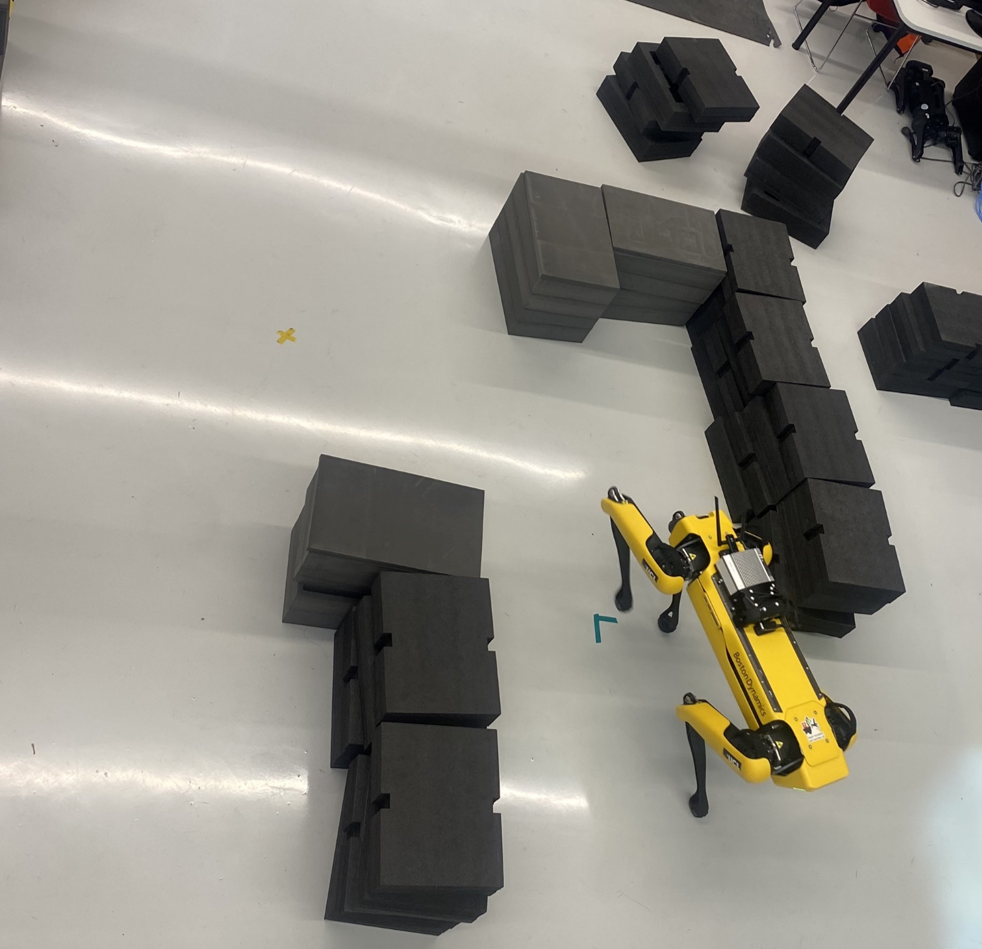}
      \label{subfig:spot_nav_around_obstacle_4}%
      }
      \vspace{0.1cm}
     \subfloat{%
      \includegraphics[height = 2.1cm]{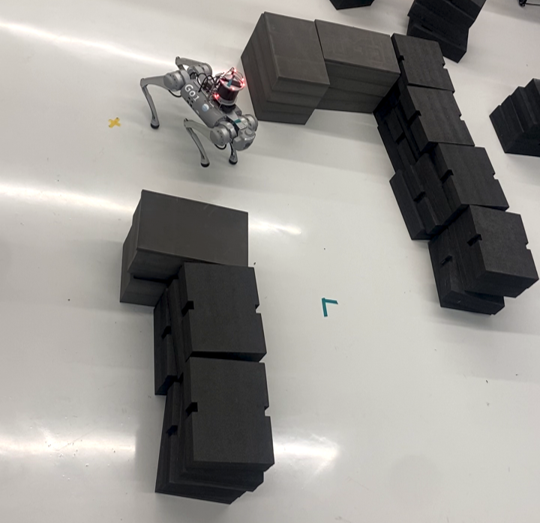}
      \label{subfig:Go1_nav_around_obstacle_1}%
      }      
    \subfloat{%
      \includegraphics[height = 2.1cm]{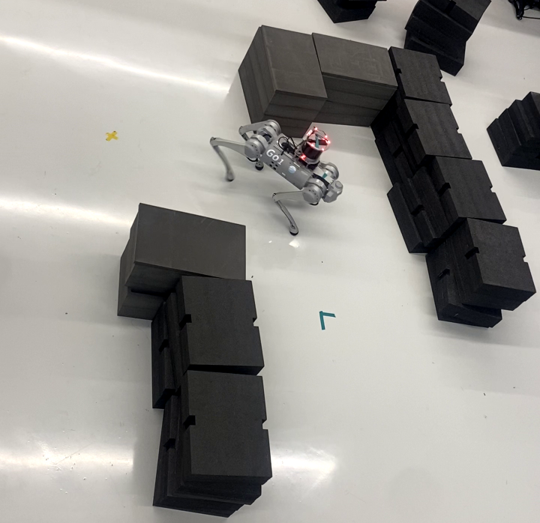}
      \label{subfig:Go1_nav_around_obstacle_2}%
      }      
    \subfloat{%
      \includegraphics[height = 2.1cm]{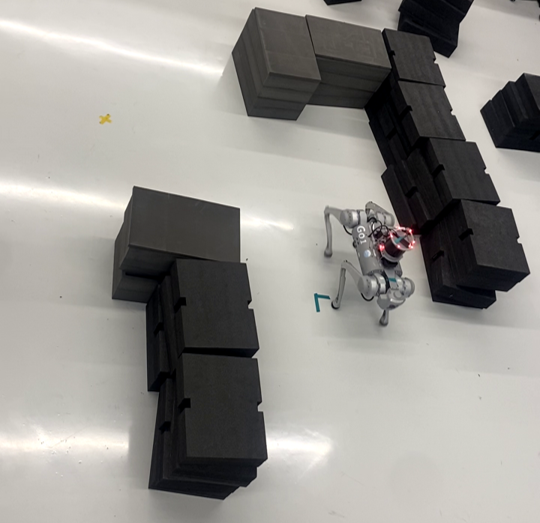}
      \label{subfig:Go1_nav_around_obstacle_3}%
      }      
    \subfloat{%
      \includegraphics[height = 2.1cm]{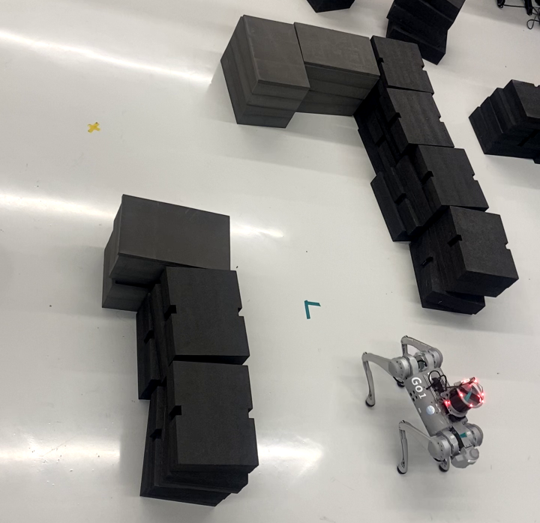}
      \label{subfig:Go1_nav_around_obstacle_4}%
      } 

    \caption{Real World Deployment: Spot (top) and Go1 (bottom) navigating around a maze environment using DiPPER in combination with the developed navigation stack (Fig.~\ref{fig:navigation_stack}).}
    \label{fig:robots_nav_around_obstacle}
\end{figure*}
The chosen  evaluation metric is the success rate. Success is achieved when the output trajectory $A_{0}$ connects the start and goal points while following a feasible path. Some examples of achieved successful trajectories are shown in Fig.~\ref{fig:resultfigures}.
We evaluate DiPPeR in 10 images from the validation set and 10 out-of-distribution maps. For each map we generate 10 random start and goal positions and repeat inference 3 times, to ensure consistency of the output. The success rate is calculated by dividing the number of successful experiments by the total number experiments and is then converted to a percent value. The start and goal position conditioned version of the DiPPeR outperformed the non-conditioned one in all tests and we set it as our default DiPPeR version. 

The percentage success rate $\%{sr}$ for DiPPeR is presented in Table~\ref{tab:successratio}.
\begin{table}[hbt!]
    \centering
    \begin{tabular}{ |c| c|} 
     \hline
        Dataset & $\%{sr}$   \\ [0.5ex] 
        \hline
        Validation  & 85  \\        
        Out-of-Distribution & 89 \\ 
        \hline
        Average & 87\\
     \hline
    \end{tabular}
    \caption{Inference success rate for validation and out-of-distribution datasets.}
    
  \label{tab:successratio}
\end{table}

\begin{figure}[hbt!]
\centering\includegraphics[width=0.75\columnwidth]{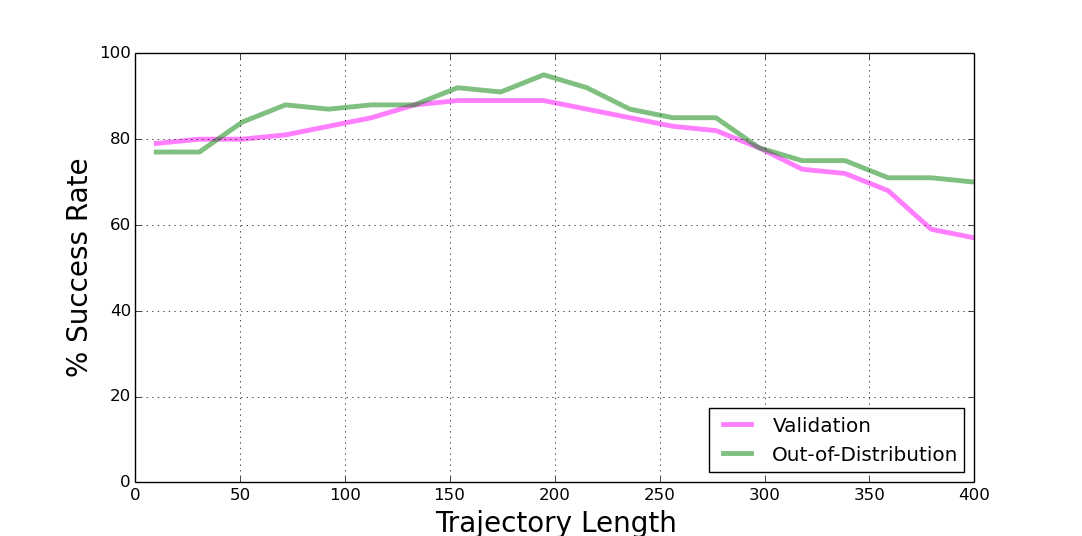}
    \caption{Plot of the  average \% success rate against the trajectory length for maps sampled from  the Validation and the Out-of-Distribution Dataset}
    \label{fig:plt_traj}
\end{figure}

DiPPeR is on average $87\%$ successful in providing feasible paths. It performs best on out-of-distribution maps as they contain instances of maps much simpler in terms obstacle structure than the training dataset. To understand the failure cases, we plot the success rate against the trajectory length (Fig.~\ref{fig:plt_traj}). The success rate drops for both extremes of smaller and larger trajectories and performs the best for trajectories of length close to $180$. This is expected as the choice of horizon length is set equal to $180$ during training, however it presents a limitation that we aim to address in future work.

We consistent the same experiments in the real-word with the  results being consistent with the simulation ones.

To assess the performance of DiPPeR against SOTA path planning frameworks, we evaluate its convergence speed against a search-based planner $A^*$ and its data driven variants N-$A^*$~\cite{yonetani2021path} and ViT-$A^*$~\cite{liu2023vit}. We compare the algorithms by using maps of increasing size and obstacle structure depicted on Fig.~\ref{fig:maps_bench}. For each map $10$ trajectories with random start and end points are generated by the $4$ algorithms. The time taken for trajectory generation is measured for all experiments and the $10$ values for each map are averaged. The results are summarized in Table~\ref{tab:sim_run_time_comparision}.

\begin{table}[hbt!]
    \centering
    \begin{tabular}{ |c| c c c c|} 
     \hline
     maps & DiPPeR &  ViT-$A^*$ & N-$A^*$ &  $A^*$ \\ [0.5ex] 
     \hline
     (a) & \textbf{0.4}   & 5.68        & {4.70}     & 6.03 \\ 
     (b) & \textbf{0.4}   & 17.31       & {14.73}    & 17.51 \\  
     (c) & \textbf{0.4}   & 4.81        & 5.17       & 15.59 \\
     (d) & \textbf{0.4}   & 12.73       & 16.57     & 36.24 \\
     \hline
    \end{tabular}
    \caption{Average time in seconds taken for trajectory generation by DiPPeR, ViT-$A^*$, N-$A^*$,  $A^*$. Maps a)-d) are presented in Fig.~\ref{fig:maps_bench}.
    \label{tab:sim_run_time_comparision}
    }
\end{table}

The maps for comparison are sampled from the MRPB benchmark dataset~\cite{wen2021mrpb}.

All experiments were conducted using a NVIDIA RTX 3090 GPU. DiPPeR is on average $23$ times faster against the next best performing SOTAs algorithms, with feasible trajectory generation taking only $0.4s$ regardless of maze size or trajectory length. This is due to the generative properties of diffusion planner.

\subsection{Real-World Deployment}
\begin{figure}[hbt!]
    \centering
    \includegraphics[width=.75\columnwidth]{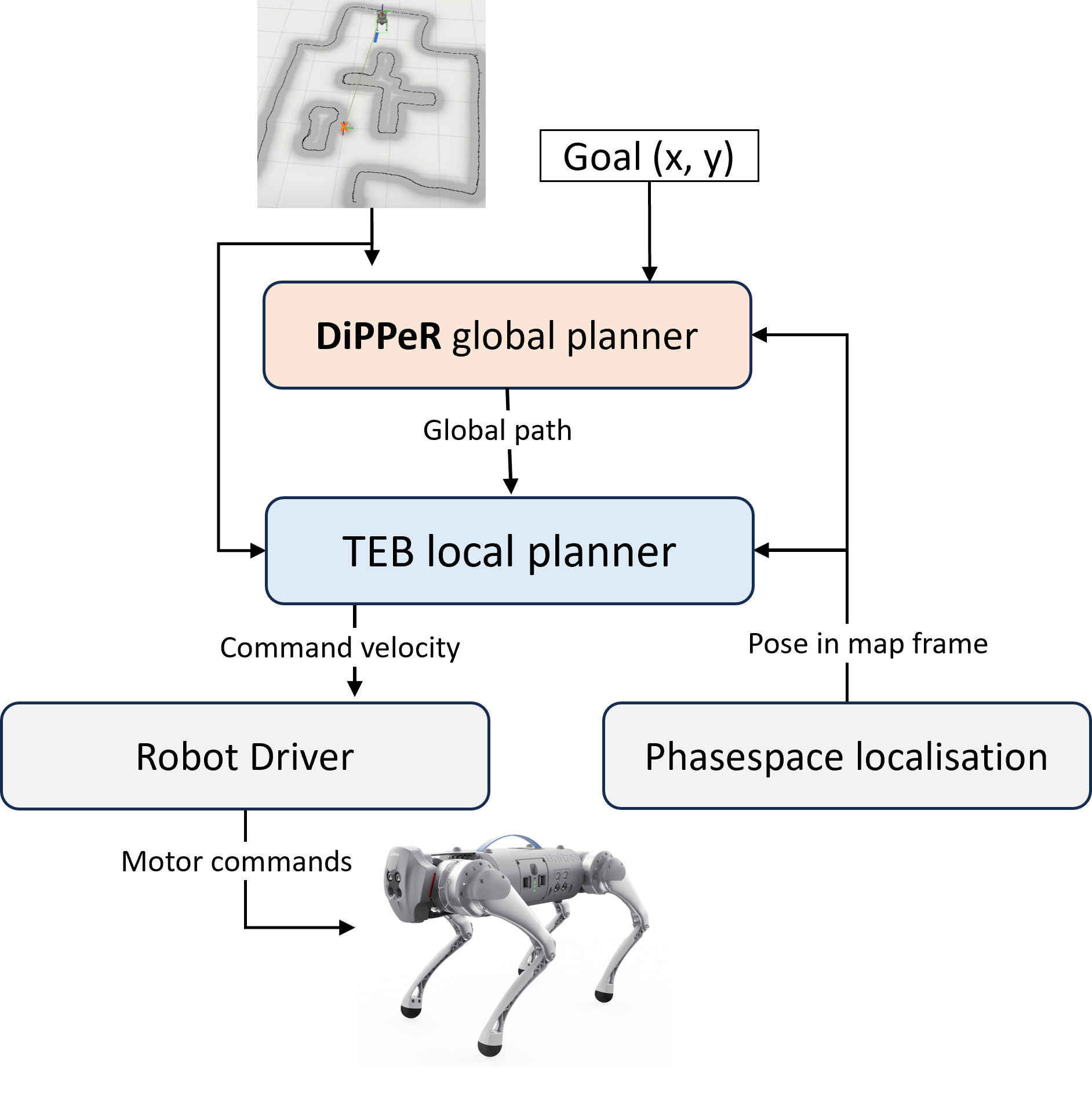}
    \caption{Schematic structure of the navigation stack for DiPPeR real robot deployment.}
    \label{fig:navigation_stack}
\end{figure}

A schematic representation of the real work deployment of DiPPeR is depicted in Fig.~\ref{fig:navigation_stack}. We validate the performance of DiPPeR in the real world and its platform agnostic property through deployment on Unitree Go1\footnote{https://www.unitree.com/en/go1/} and Boston Dynamics Spot\footnote{https://www.bostondynamics.com/products/spot}.

In order to leverage the existing robot navigation frameworks, DiPPeR is integrated with the 2D ROS navigation Stack\footnote{http://wiki.ros.org/navigation}, to act as a global path planner. Given the occupancy map, DiPPeR generates a global path which is further refined by the local planner -- to avoid violation of the robots kinodyanmics constraints and an external tracking system -- to mitigate for state estimation inaccuracies. 
The local planner used is the Timed-Elastic-Band (TEB)~\cite{rosmann2012trajectory, rosmann2013efficient} and the external tracker of choice is Phasespace tracking cameras\footnote{https://www.phasespace.com/}. Phasespace cameras allow for 960 Hz robot real-time localization. Examples of successful deployment of the pipeline can be seen in Fig.~\ref{fig:robots_nav_around_obstacle}.

\section{CONCLUSION}\label{sec:conclusion}
In this paper we present DiPPeR, an image guided diffusion based 2D-path planner. The planner is successful in generating feasible paths of variable length on average $80\%$ of the time, for maps of various size and obstacle structure. DiPPeR outperformed in speed against both search based and data driven planner by a factor of $23$. We validate the transfer of the planner in the real world and showcase its platform agnostic capabilities by successfully testing it on two different robots. 
We identified that DiPPeR tends to performs less optimally for trajectories significantly longer than the training horizon, as well as requiring an estimate of the number of step for the final trajectory during inference. We are aiming to address this in future work by also experimenting with different network architectures, such as transformers, that show promising results in the field of diffusion.
% \section{ACKNOWLEDGMENTS}
% We would like to thank Davide Paglieri for sharing his insights regrading the diffusion pipeline training. 
\clearpage

% \section*{Acknowledgments}

\addtolength{\textheight}{0cm}   % This command serves to balance the column lengths

%%%%%%%%%%%%%%%%%%%%%%%%%%%%%%%%%%%%%%%%%%%%%%%%%%%%%%%%%%%%%%%%%%%%%%%%%%%%%%%%
\bibliographystyle{IEEEtran}
\bibliography{IEEEabrv, icra_2024}
\end{document}